\documentclass[sigconf]{acmart}

\usepackage[linesnumbered,ruled,vlined]{algorithm2e} 

\DeclareMathDelimiter{(}{\mathopen} {operators}{"28}{largesymbols}{"00}
\DeclareMathDelimiter{)}{\mathclose}{operators}{"29}{largesymbols}{"01}
\AtBeginDocument{%
  \providecommand\BibTeX{{%
    \normalfont B\kern-0.5em{\scshape i\kern-0.25em b}\kern-0.8em\TeX}}}

\setcopyright{rightsretained}
\copyrightyear{2023}
\acmYear{2023}
\acmPrice{0}
\acmDOI{X}

\acmConference[HRI 2023 workshop on Variable Autonomy for human-robot Teaming]
{ACM/IEEE HRI 2023 workshop on Variable Autonomy for human-robot Teaming}
{March 13, 2023}
{Stockholm, Sweden}
\begin{document}

\title{Dynamic Competency Self-Assessment for Autonomous Agents}

\author{Nicholas Conlon}
\email{nicholas.conlon@colorado.edu}
\orcid{0000-0001-8262-2169}
\affiliation{%
  \institution{University of Colorado at Boulder}
  \streetaddress{3775 Discovery Drive}
  \city{Boulder}
  \state{Colorado}
  \country{USA}
  \postcode{80303}
}

\author{Nisar R. Ahmed}
\email{nisar.ahmed@colorado.edu}
\orcid{0000-0002-7555-5671}
\affiliation{%
  \institution{University of Colorado at Boulder}
  \streetaddress{3775 Discovery Drive}
  \city{Boulder}
  \state{Colorado}
  \country{USA}
  \postcode{80303}
}

\author{Daniel Szafir}
\email{daniel.szafir@cs.unc.edu}
\orcid{0000-0003-1848-7884}
\affiliation{%
  \institution{University of North Carolina at Chapel Hill}
  \streetaddress{201 S. Columbia Street}
  \city{Chapel Hill}
  \state{North Carolina}
  \country{USA}
  \postcode{27599}
}

\renewcommand{\shortauthors}{Conlon, et al.}
\newcommand{\nickComm}[1]{{\color{red}{ \textbf{NJC:} #1}}}
\begin{abstract}
As autonomous robots are deployed in increasingly complex environments, platform degradation, environmental uncertainties, and deviations from validated operation conditions can make it difficult for human partners to understand robot capabilities and limitations. The ability for a robot to self-assess its competency in dynamic and uncertain environments will be a crucial next step in successful human-robot teaming. This work presents and evaluates an Event-Triggered Generalized Outcome Assessment (ET-GOA) algorithm for autonomous agents to dynamically assess task confidence during execution. The algorithm uses a fast online statistical test of the agent's observations and its model predictions to decide when competency assessment is needed. We provide experimental results using ET-GOA to generate competency reports during a simulated delivery task and suggest future research directions for self-assessing agents.
\end{abstract}





\received{1 March 2023}

\maketitle

%%%%%%%%%%%%%%%%%%%%%%%%%%%%%%%%%%%%%%%%%%%%%%%%%%%%%%%%%%%%%%%%%%%%%%%%%%%%%%%%
\section{INTRODUCTION} \label{sec:intro}
Autonomous self-assessments are a critical component to advancing complex robot deployments. Consider a scenario where a search-and-rescue (SAR) team is delivering much needed supplies after a disaster. The environment is quite dangerous so the team decides to employ semi-autonomous robots to navigate the environment and deliver the supplies. The team's reliance on the robot is based on their perception of its ability. However, if there is misalignment between the team's perception of the robot's abilities, and the robot's actual capabilities and limitations, the team may inadvertently push the robot beyond its competency boundaries \cite{Hutchins2015}. To make appropriate tasking decisions (e.g., safe delivery locations, control handoffs, choice of autonomy level), the SAR team must understand the robot's competency and how it may change during the mission. 

In recent work, we showed that agents which reported \textit{a priori} self-assessed confidence in task success helped align operator perception with actual robot competency, thus improving decision making and performance in a variable-autonomy navigation task \cite{conlon2022_iros}. However, in dynamic and uncertain environments like the SAR scenario outlined above, an \textit{a priori} confidence assessment can quickly become stale due to environmental changes like falling debris or obstacles. In this work, we propose an algorithm called \textit{Event-Triggered Generalized Outcome Assessment} (ET-GOA) which enables autonomous agents to continually monitor \textit{in situ} changes to competency, and if necessary, self-assess and report in response to events that significantly alter outcome likelihoods. We demonstrate that our method can capture dynamic environmental changes that positively or negatively impact task success and that using ET-GOA confidence assessments in autonomous decision-making can lead to significant increases in task performance.

%%%%%%%%%%%%%%%%%%%%%%%%%%%%%%%%%%%%%%%%%%%%%%%%%%%%%%%%%%%%%%%%%%%%%%%%%%%%%%%%
\section{Background and Related Work}
Competency self-assessment enables autonomous agents to assess their capabilities and limitations with respect to task constraints and environmental conditions. This critical information can be used to improve internal decision-making and/or can be communicated to a human partner to improve external decision-making. Pre-mission (\textit{a priori}) self-assessments enable an autonomous agent to assess its competency before execution of a task or mission. These methods generally compute agent self-confidence based on simulation \cite{ardon2020, israelsen2019} or previous experience \cite{frasca2020}. Our recent work showed that reporting of \textit{a priori} self-assessments lead to better choices of reliance \cite{conlon2022_psa} and improvements to performance and trust \cite{conlon2022_iros}. However in dynamic environments, \textit{a priori} assessment is a poor predictor of the agent's confidence due to factors that are not accounted for before execution, such as environmental changes, task changes, or interactions with other agents. Running \textit{a priori} methods online (periodically) could conceivably capture dynamic competency changes. However such assessments can waste computational resources if competency has, in fact not changed, or may be too expensive for certain kinds of decision-making agents \cite{conlon2022_caml, acharya2022competency, gautam2022_aat}. 

In-mission (\textit{in situ}) self-assessment enables an autonomous agent to assess (or reassess) its competency during task execution. Popular methods such as online behavior classification can identify poor behavior and trigger the agent to halt operation and ask for help in the event of a failure \cite{rojas2017,FOX2006, wu2017}. These methods, while able to capture dynamic competency changes, require examples of both good (competent) and poor (incompetent) behavior, which may be difficult or impossible to acquire in many real-world applications. Another method of \textit{in situ} self-assessment involves monitoring features of the agent's current state. For example, Gautam et al. developed a method to monitor deviations from design assumptions \cite{gautam2022_aat}, while Ramesh et al. used the ``vitals'' of a robot to monitor its health during task execution \cite{ramesh2022_vitals}. Both methods provide a valuable instantaneous snapshot of the agent's state at a given time which can indicate performance degradation online; however, neither predicts higher level task competency (e.g., does the degradation actually impact the task outcome?). In contrast, we propose a method of \textit{in situ} self-assessment that offers a hybrid approach by monitoring the alignment between the agent's predictions and observations and triggering a (re)assessment of task confidence using an accurate \textit{a priori} method when there is a deviation in alignment.

%%%%%%%%%%%%%%%%%%%%%%%%%%%%%%%%%%%%%%%%%%%%%%%%%%%%%%%%%%%%%%%%%%%%%%%%%%%%%%%%
\section{Dynamic Self-Assessment of Task Confidence}

\subsection{When to Assess Competency}
To prevent unnecessary assessments and save onboard computational resources, we take an event triggered approach to \textit{in situ} self-assessment. Our algorithm assesses confidence only when there is evidence that the agent's task confidence has changed.

One promising method for detecting such a change is the Surprise Index (SI). SI is defined as the sum of probabilities of more extreme (or less probable) events than an observed event given a probabilistic model \cite{Zagorecki2015}. For a given event $e \in E$, SI is computed by summing over the probabilities of more extreme events in the distribution $p(E)$:
\begin{align}
  SI(e, p(E))=\int_{p(E)<p(e)} p(E) dE
\end{align}
SI can be though of as how (in)compatible an observation $e$ is given a set of possible events $E$. This is a similar to the more well known entropy based surprise \cite{baldi_bitswows, benish_entropy}; however entropy based surprise is unbounded, while Surprise Index is bounded between zero (most surprising) and one (least surprising). SI also shares similarities with the tail probability or the p-value given the hypothesis that $e$ is from the distribution $p(E)$; a large p-value (large SI) indicates strong evidence that $e$ is likely from $p(E)$, while a small p-value indicates strong evidence to the contrary.

In this work we are interested in determining when the agent should re-assess its task confidence. We propose computing the SI of the agent's observed state $s_t$ with respect to that agent's model prediction $p(\hat{s}_t)$, and triggering a re-assessment when the SI falls below a threshold $\delta$. In essence, we are monitoring the quality of the agent's model given the task and triggering an assessment when quality wanes. Because some aspects of $s_t$ may be more relevant to competency than others, we compute SI over a subset of the state marginals.

\subsection{How to Assess Competency}
To assess competency we leverage the Generalized Outcome Assessment (GOA) \cite{conlon2022_caml}. Given a probabilistic world model $M$, GOA simulates task execution by rolling out state predictions $p_{M}(s_{t+1}|s_{t}, a_{t})$. Note that a $M$ could take the form of a Monte Carlo based planner\cite{israelsenThesis}, a black box neural network \cite{conlon2022_caml, ha_worldmodels}, or similar. GOA then analyses the state predictions and computes the agent's margin of confidence in attaining an outcome better than some target outcome threshold $Z$. Examples of target outcomes could include craters hit (which we prefer less of) or packages delivered (which we prefer more of). The confidence value can be reported as a raw value $\in (0,1)$ or mapped to a semantic label indicating confidence such as \textit{highly likely, likely, unlikely, highly unlikely}. For the experiments outlined later, we use the raw numerical values of confidence.

\subsection{Event-Triggered Generalized Outcome Assessment Algorithm} \label{sec:ET-GOA}
We call our method for surprise-based dynamic self-assessment \textit{Event-Triggered Generalized Outcome Assessment} (ET-GOA). The algorithm is presented in Alg. \ref{alg:etgoa} and can be broken up into two components: (1) before task execution and (2) during task execution.

\noindent \textit{Before task execution (lines 1-5)}: Line 1 takes as input a model $\textbf{M}$, a task specification $\textbf{T}$, a set of outcome thresholds $\textbf{Z}$ (one for each outcome), and a set surprise thresholds $\delta$ (one for each state marginal of interest). Next (line 2) the model $\textbf{M}$ is used to simulate execution of task $\textbf{T}$ given initial state $s_0$. This results in a set predicted state distributions $[p(\hat{s}_t)]_{t=0:N}$, one for each time step $t$. We further break the state distribution for a given time step into $K$ marginal components. For example if we were interested in using the $x, y,$ and $z$ position in the SI trigger, $K=3$ and $p(\hat{s}_t)$ would be broken into the set of marginal probability distributions $[p(\hat{s}_{t,x}), p(\hat{s}_{t,y}), p(\hat{s}_{t,z})]$. This additional step of marginalization is implicit in the algorithm, but important to note. The predicted marginals for each time step are then stored in an experience buffer (line 3), and then used to compute the initial Generalized Outcome Assessment (line 4), which can be reported to an operator (line 5). 

\begin{algorithm}
  \DontPrintSemicolon
  \SetKwFunction{FMain}{ET-GOA}
  \SetKwProg{Fn}{Algorithm}{:}{}
  \SetKwFor{For}{for (}{) }{}
  \Fn{\FMain{$\textbf{M}$, $\textit{T}$, $\textbf{Z}$, $\delta$}}{
        $[p(\hat{s}_1),...,p(\hat{s}_N)] \gets$ simulate $\textbf{M}(\textit{T}, s_0)$ \;
        
        exp\_buffer $\gets [p(\hat{s}_1),...,p(\hat{s}_N)]$\;
        $goa \gets$ GOA(exp\_buffer, $\textbf{Z}$)\;
        report goa\;
        \For{$t$ in $1:N$}{
            $s_{t} \gets$ receive\_state\_observation($t$)\;
            $p(\hat{s}_{t})$ $\gets$ exp\_buffer($t$)\;
            $si_{min} = \min_{i=1:K}{\textit{SI}(s_{t,i}, p(\hat{s}_{t,i}))}$\;
            \If{$si_{min}\leq \delta$}{
                $[p(\hat{s}_{t+1}),...,p(\hat{s}_N)] \gets$ simulate $\textbf{M}(\textit{T}, s_t)$ \;
                exp\_buffer $\gets [p(\hat{s}_{t+1}),...,p(\hat{s}_N)]$\;
                $goa \gets$ GOA(exp\_buffer, $\textbf{Z}$)\;
                report goa\;
            }
            \Else {
            continue\;
            }
            
         }
  }
  \caption{Event-Triggered Generalized Outcome Assessment}
  \label{alg:etgoa}
\end{algorithm}

\noindent \textit{During task execution (lines 6-16)}: The agent  observes the state $s_t$ at time $t$ (line 7). It then retrieves the state distributions (i.e., the predictions) for time $t$ from the experience buffer (line 8). Next the algorithm computes $si_{min}$, the minimum of the SI of each of the $K$ observed state marginals $s_{t,i}$ given the predicted marginal distributions $p(\hat{s}_{t,i})$ (line 9). If $si_{min}$ is below $\delta$ an anomalous or surprising state observation has been received and confidence should be reassessed (line 10). In this case, a new set of predicted state distributions are simulated from $M$ (line 11) and saved in the experience buffer (line 12). A new self-assessment is then computed using the newly updated experience buffer (line 13) and reported to an operator (line 14). $si_{min}$ above $\delta$ indicates that the the agent's predictions align with its observations and no confidence update is needed at this time (line 16). This loop (line 6) continues for the duration of the task, comparing predicted state marginal distributions to real observations and (if necessary)  recomputing and reporting updates to the agent's task confidence.

%%%%%%%%%%%%%%%%%%%%%%%%%%%%%%%%%%%%%%%%%%%%%%%%%%%%%%%%%%%%%%%%%%%%%%%%%%%%%%%%
\section{Experiments}
We evaluated ET-GOA in two simulation experiments. The first investigated ET-GOA's impact on task performance. The second investigated ET-GOA's ability to capture changes in task difficulty.

\subsection{Delivery Scenario Overview}
Our experimental scenario was based on the motivating SAR example from section \ref{sec:intro}: A single agent was tasked to safely deliver cargo to one of three goals. The environment contained two types of obstacles: craters and dust zones, which were difficult for the agent to avoid. Driving over craters damaged the agent, and if enough craters were hit while navigating then the agent was considered broken and failed the delivery task. Dust zones degraded sensors and injected noise into the agent's state transition dynamics. Dust zones were generally found near craters, which increases the chance that the agent hit a craters if it found itself in dust. To simulate environmental changes that would occur in realistic deployments, new obstacles could spawn at random locations (except for the agent's location) during task execution. 

The environment was a custom OpenAI Gym environment \cite{brockman2016openaigym}. The agent was modeled as a discrete state/action Markov Decision Process with state space $s=(s_x,s_y,s_c,s_z)$ consisting of the agent's $(x,y)$ location and the counts of craters $(s_c)$ and dust zones $(s_z)$ within its sensor field of view (FOV). The sensor FOV was modeled as omnidirectional with a radius of 10 grid squares. The total size of the 2D environment was 50x50 grid squares. We trained one policy for each goal using Q-Learning \cite{watkins1992q}. No obstacles were present during training to prevent the agent from learning how to overcome the difficulties of the environment. 

\begin{figure}[!htbp]
    \centering
    \includegraphics[width=0.45\textwidth]{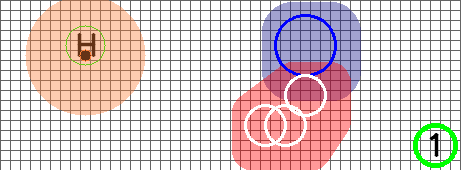}
    \caption{Example environment illustrating the agent's location and FOV (orange), the goal area (green), truth locations of dust zones (blue circle) and craters (white circles). The obstacles are highlighted with blue and red to improve visual contrast.}
    \label{fig:env}
\end{figure}

The world model $M$ used for self-assessment was a copy of the environment that had an identical state transition function but only included known craters and dust zones. The agent chose the goal which had the maximum assessed confidence. If there was a tie in confidence, the agent chose the closest goal. An example environment can be seen in Fig. \ref{fig:env}.

We evaluated two different environments, \textit{static} and \textit{dynamic}. In the static environment, the locations of craters and dust zones were known by the agent \textit{a priori} and remained unchanged for the entire task execution. In the dynamic environment, the locations of craters and dust zones were initially known, but changed at a predetermined time to simulate a previously generated onboard navigation map suddenly becoming out-of-date. 

\subsection{Hypotheses}
We had three hypotheses: (1) In a static environment, ET-GOA and GOA will perform equally well and will both outperform a random goal choice; (2) In a dynamic environment, ET-GOA will outperform both GOA and random choice; (3) ET-GOA can capture both positive and negative changes in task difficulty. We analyzed agent performance (number of deliveries) for hypothesis 1 and 2, and we analyzed reported confidence relative to task difficulty changes for hypothesis 3.

\subsection{Improvements to Performance}
Our first experiment was used to validate our first two hypotheses. At $t=0$ we initialized the agent with the locations of all obstacles. For dynamic conditions we changed the locations of the obstacles without the agent's knowledge at $t=10$. Three conditions were considered: \textit{no assessment}, \textit{GOA}, and \textit{ET-GOA}. The \textit{no assessment} condition did not use any competency assessment. Rather, at $t=0$ the agent chose the goal at random and navigated directly to it. The \textit{GOA} condition used the standard Generalized Outcome Assessment analysis discussed in \cite{conlon2022_caml}. At $t=0$ the agent selected and navigated to the goal $g\in G$ with the highest GOA confidence according to Eqn. \ref{eqn:argmax_goal}.
\begin{equation}
    g = \arg \max_{i\in G}{GOA_i}
    \label{eqn:argmax_goal}
\end{equation}

The \textit{ET-GOA} condition used the ET-GOA algorithm discussed in \ref{sec:ET-GOA}. We used two state marginals as triggers: $s_c$ (and $\hat{s}_c$), the actual (and predicted) craters visible in the agent's FOV; and $s_z$ (and $\hat{s}_z$), the actual (and predicted) dust zones visible in the agent's FOV. This essentially computed the surprise between the expected obstacle locations (from the initial location information) and the ``on the ground'' obstacle locations observed while traversing the environment. We chose these specific marginals because they align with the sensor capability in modern robots. Additionally, looking at surprising observations in the agent's sensor FOV enabled the algorithm to trigger a re-assessment prior to the agent physically coming into contact with a possibly dangerous obstacle. The algorithm triggered a re-assessment if the minimum SI of either state marginal was less than $\delta=0.05$:
\[
\min(SI(s_{c},\hat{s}_c), SI(s_{z},\hat{s}_z)) < 0.05
\]
The agent then selected the goal based on eqn. \ref{eqn:argmax_goal}. The agent navigated directly to that goal until it either reached the goal or triggered a re-assessment and chose a new goal. For each condition the agent attempted 100 delivery tasks.

\subsubsection{Results}
\begin{figure}[!htbp]
    \centering
    \includegraphics[width=0.45\textwidth]{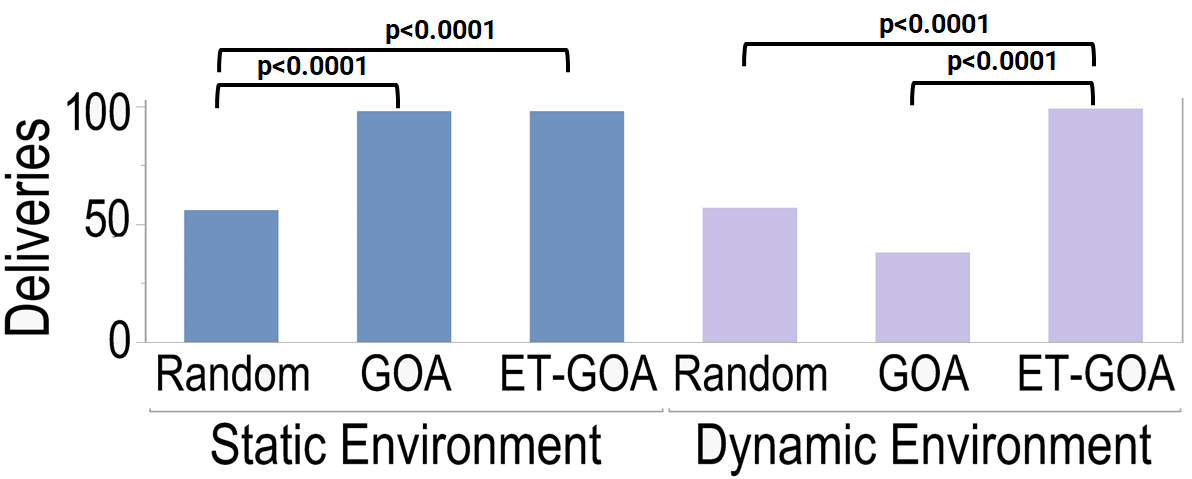}
    \caption{Plot of 100 delivery attempts per condition showing ET-GOA preformed significantly better than random in the static environment (left) and significantly better than both random and GOA in the dynamic environment (right).}
    \label{fig:experiment_1}
\end{figure}

We found significant main effects of the environment on number of deliveries ($t(598)=4.65, p<0.0001$) indicating that the static environment was easier than the dynamic environment, which was expected. In the static environment, we found significant effects of each reporting condition on deliveries ($F$($2,297$)$=43.5, p<0.0001$). Post-hoc analysis using Tukey's HSD revealed significant increase in deliveries in ET-GOA compared to random ($p=0.0001$), and significant increase in deliveries in GOA compared to and random ($p<0.0001$). There was no difference between GOA and ET-GOA in the static environment, which was expected. In the dynamic environment, we found significant effects of each reporting condition on deliveries ($F$($2,297$)$=44.1, p<0.0001$). Post-hoc analysis using Tukey's HSD revealed significant increase in deliveries in ET-GOA compared to both random ($p=0.0001$) and GOA ($p<0.0001$). These results confirm our first and second hypotheses and can be seen in Fig. \ref{fig:experiment_1}.

\subsection{Detecting Changes in Difficulty}
Our second experiment was used to validate our third hypothesis. Here we evaluated at how well ET-GOA captured changes in the environment that impacted task difficulty. For this evaluation, we isolated the agent to navigate to a single static goal location under two conditions. The first, called $easy\xrightarrow{}hard\xrightarrow{}easy$, starts out with no obstacles, then obstacles are randomly added at time step 10, then all obstacles are deleted at time step 30. The second, called $hard\xrightarrow{}easy\xrightarrow{}hard$, starts out with randomized obstacles, then all obstacles are deleted at time step 10, and then new obstacles are added at time step 30. Adding obstacles increases task difficulty for the agent and \textit{vice versa} for deleting obstacles. Obstacles at time step zero were known to the agent, while the obstacles added/deleted at time steps 10 and 30 had to be observed \textit{in situ}. We ran 100 episodes for each condition and recorded the initial assessment and the ET-GOA assessment after each add/delete event.

\subsubsection{Results}
\begin{figure}[!htbp]
    \centering
    \includegraphics[width=0.5\textwidth]{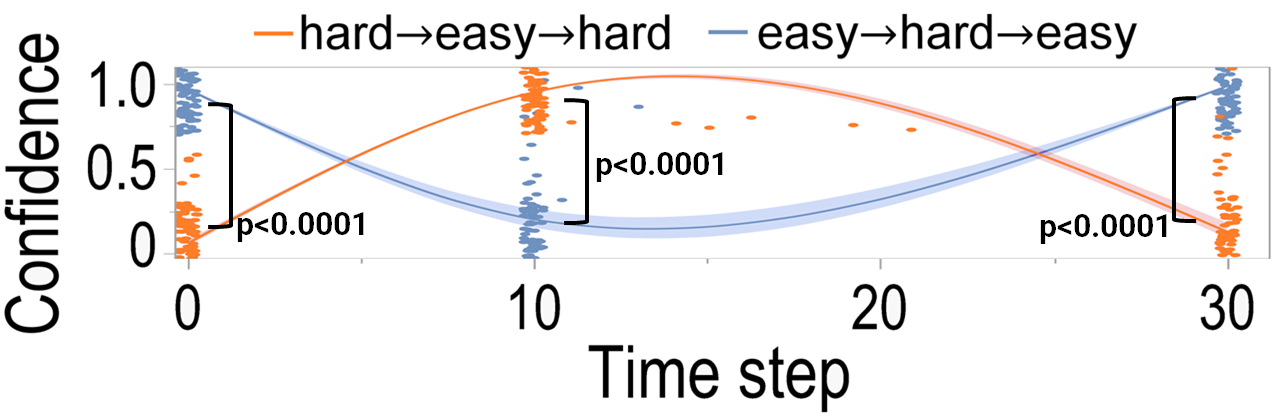}
    \caption{Plot showing ET-GOA captured task difficulty changes. The $hard\xrightarrow{}easy\xrightarrow{}hard$ tasks are in orange, while $easy\xrightarrow{}hard\xrightarrow{}easy$ tasks are in blue. Task difficulty changing events occurred at $t=10$ and $t=30$. The solid lines indicate task confidence mean and standard deviation.}
    \label{fig:difficulty}
\end{figure}

We observed a significant difference in the agent's confidence between $easy\xrightarrow{}hard\xrightarrow{}easy$ tasks and $hard\xrightarrow{}easy\xrightarrow{}hard$ at the initial assessment $(t(99)=110.0, p<0.0001)$, after the first environmental change $(t(99)=27.7, p<0.0001)$, and after the second environmental change $(t(99)=37.5, p<0.0001)$. This confirms our third hypothesis that ET-GOA can capture both positive and negative impacts to task difficulty. A plot of the results can be seen in Fig. \ref{fig:difficulty}.

%%%%%%%%%%%%%%%%%%%%%%%%%%%%%%%%%%%%%%%%%%%%%%%%%%%%%%%%%%%%%%%%%%%%%%%%%%%%%%%%
\section{Conclusions and Future Work}
In this work we presented an algorithm for \textit{Event-Triggered Generalized Outcome Assessment} which computes an autonomous agent's \textit{in situ} task confidence in dynamic and uncertain environments. ET-GOA chooses when to assess task confidence based on the Surprise Index between an agent's predicted and actual state. We evaluated ET-GOA on a delivery task in both static and dynamic environments and found that it led to significant performance improvements over baseline methods. We also found that ET-GOA was able to capture changes in agent confidence indicating changes in task difficulty. That is, our method can determine when tasks become more or less difficult. Our next step is to validate ET-GOA both on live platforms and in a human subjects study. We hypothesize that the presence of ET-GOA will help operators make better decisions when it comes to relying on an autonomous robot, leading to improved performance and reductions in workload.

ET-GOA can enable autonomous robots to provide critical information about their ``on the ground'' confidence in task success, when that confidence changes, and why. We believe that it can be invaluable to human-robot teams, particularly those working in high risk and uncertain environments where human operators need to make critical decisions with respect to task execution, level of autonomy, and/or control.

\begin{acks}
This work was supported by the Defense Advanced Research Projects Agency (DARPA) under Contract No. HR001120C0032. Any opinions, findings and conclusions or recommendations expressed in this material are those of the author(s) and do not necessarily reflect the views of DARPA.
\end{acks}

\bibliographystyle{ACM-Reference-Format}
\bibliography{refs}

\end{document}